\documentclass[10p,times]{elsarticle}

\usepackage{amssymb}
\usepackage{amsmath}
\usepackage{graphicx}
\usepackage{orcidlink}
\usepackage{here}
\usepackage{natbib}
\usepackage{algorithm}
\usepackage{xcolor}

\begin{document}

\journal{Expert Systems with Applications}

\begin{frontmatter}

\title{Meta-classification of one-class classification models using ranking correlation and nearest neighbor}

\author[a]{Toshitaka Hayashi \orcidlink{0000-0002-7599-4404}}
\ead{toshitaka.hayashi@uhk.cz}
\author[a,b,c]{Hamido Fujita\corref{cor1} \orcidlink{0000-0001-5256-210X}}
\ead{fujitahamido@utm.my, HFujita-799@acm.org}
\author[a]{Dalibor Cimr \orcidlink{0000-0003-2197-8553}}
\ead{dalibor.cimr@uhk.cz}
\author[a]{Richard Cimler \orcidlink{0000-0001-6712-9894}}
\ead{richard.cimler@uhk.cz}
\author[a]{Jitka Kühnová \orcidlink{0000-0001-9223-5672}}
\ead{jitka.kuhnova.2@uhk.cz}
\cortext[cor1]{Corresponding author: Professor Hamido Fujita \\ Email: fujitahamido@utm.my, HFujita-799@acm.org \\  Postal address: Kotorizawa, 2-27-5, Morioka, Iwate 020-0104, Japan\\  Tel: +81 8067208218}
\affiliation[a]{organization={Faculty of Science, University of Hradec Kralove}, addressline={Hradecká 1285}, city={Hradec Kralove}, postcode={50003}, country={Czech Republic}}
\affiliation[b]{organization={Malaysia-Japan International Institute of Technology (MJIIT), Universiti Teknologi Malaysia}, addressline={Jalan Sultan Yahya Petra, Kampung Datuk Keramat}, city={Kuala Lumpur}, postcode={54100}, country={Malaysia}}
\affiliation[c]{organization={Regional Research Center, Iwate Prefectural University}, addressline={Sugo 152-52}, city={Takizawa}, postcode={020-0693}, state={Iwate}, country={Japan}}

\begin{abstract}
Machine Learning (ML) techniques have been applied to various problems. However, applying ML to ML models is an unexplored direction. For this purpose, this paper considers a meta-classification of one-class classification (OCC) models, because all ML models could be approximated as OCC models. The proposal represents OCC models as normality rankings and classifies them using nearest-neighbor and ranking-correlation metrics. The experiment classifies OCC models, where classes correspond to training datasets, algorithms, and hyperparameters. The proposal achieves high accuracy when class labels are datasets. Moreover, it can classify algorithms when the training datasets contain the same class. In addition, the discussion highlights that the classification of OCC models is essentially the classification of datasets that treats multiple samples as a single input. The experiment demonstrates the classification of datasets using sleeping records. The proposed method can provide a unified solution for classifying OCC models, datasets, and rankings. Source code is uploaded to the public repository https://github.com/ToshiHayashi/ClassOCC.
\end{abstract}


\end{frontmatter}
\section{Introduction}
Machine learning (ML) has been applied to several problems \citep{[17],[18],[19]}. However, ML was not applied to ML models themselves. The goal of this research is to address the question: how to apply ML to ML models. However, "applying ML to ML" is a vague topic because ML has several tasks, such as classification, clustering, and regression. To simplify the discussion, this paper focuses on the classification of ML models. 

Subsequently, there is a question about what ML models are classification targets. Ultimately, all ML models should be classified. However, some generalization is required as ML addresses a wide range of tasks. 

This paper selects one-class classification (OCC) models as classification targets. OCC is a classification problem in which the training set consists solely of one class \citep {OCCreview}. OCC models are ideal classification targets because they can learn both labeled and unlabeled datasets. Moreover, transforming other ML models into OCC models is relatively easy. For instance, binary or multi-class classification models can be approximated as a set of OCC models \citep{decompose}. In addition, the subtask-based OCC approach\citep{OCCreview} utilized errors of other ML models. In other words, one can approximate most ML models as the subtask-based OCC models. 

Accordingly, this study aims to address the question how to classify OCC models. Generally, OCC models contain classification functions, score functions, and threshold values. Analyzing score functions is reasonable because they are the most unique part of OCC models. However, direct comparison of score functions is not possible because OCC algorithms have individual score functions; for instance, one algorithm uses some variables, while others do not. 

To address this issue, this study analyzes model outputs for common inputs. In particular, the proposed method constructs normality rankings from score functions and classifies the rankings using nearest-neighbor \citep{NN} and ranking correlation \citep{[2],[3]}. 

This paper is an extension of a conference paper \citep{[1]} that applied anomaly detection (AD) to AD models. However, the concept "AD" was vague; the actual content focused on the classification of OCC models and binary classification models. In addition, the binary classification models were classified as the probability of the single class. Accordingly, this paper reorganizes the contents as a classification of OCC models. Moreover, there is a preprint \citep{KEnsemble} that applies clustering to OCC models. The preprint used the same OCC models as this paper.

The main change from the conference paper \citep{[1]} is to include more detailed experiments, such as 1) classification of ensemble models, 2) classification of data balance, and 3) classification of hyperparameters. Moreover, the journal version highlights the function of classifying datasets. OCC models and training datasets have one-to-one correspondence. Therefore, the classification of OCC models is essentially the classification of datasets. This paper demonstrates the above function using sleep records. 

The contribution is listed as follows:
\begin{itemize}
\item The main originality is the classification of OCC models. To the best of our knowledge, this is the first journal article to classify ML models. The paper also provides theoretical insights into the classifications of datasets and rankings. 
\item The proposed method is a unified solution for the classification of OCC models, datasets, and rankings.
\item The experiment classifies OCC models in terms of the training datasets, algorithms, and hyperparameters.
\item The discussion section demonstrates the proposal using a biometric signal and sleep records.
\end{itemize}

The remainder of the paper is organized as follows. Section 2 summarizes related work. Section 3 proposes a classification algorithm for OCC models. Sections 4 and 5 are the experiment and discussion, respectively. Finally, Section 6 is a conclusion.

\section{Related Work}
\label{sec2}
This section summarizes related work. 
Subsection \ref{sec2.1} describes one-class classification. Subsection \ref{sec2.2} mentions meta-learning and explainable artificial intelligence. Subsection \ref{sec2.3} informs the ranking correlation.

\subsection{One-class classification}
\label{sec2.1}
This paper defines OCC, OCC algorithms, and OCC models as follows:
\begin{itemize}
\item \textbf{OCC} is a classification problem where the training dataset contains solely one class. This definition is rigor.
\item \textbf{OCC algorithms} refer to the algorithms that "can" train the model from a single class. One can apply OCC algorithm to multiple classes by approximating that all training samples belong to a single class. 
\item \textbf{OCC models} refer to the models trained by OCC algorithms. The models might learn multiple classes.
\end{itemize}
This paper aims to classify OCC models. However, the task itself is not OCC if multiple classes are involved to train the models.

Generally, the OCC algorithm trains the model from a dataset.
\begin{equation}\label{eq1}
Training: dataset \rightarrow model
\end{equation}

The OCC model contains a classification function, a score function, and a threshold value $\lambda$.
The classification function classifies samples:
\begin{equation}\label{eq2}
Classification: X \rightarrow class
\end{equation}
where classes include one known class and the remaining unknown classes.

The general classification function is written as:
\begin{equation}\label{eq3}
Classification(X)=\left\{\begin{alignedat}{2}
One (score(X)\ge\lambda) \\
Other (score(X)< \lambda)
\end{alignedat}
\right.,
\end{equation}
where the score function computes the normality of the sample:
\begin{equation}\label{eq4}
score: X \rightarrow normality
\end{equation}
Normality refers to the likelihood of belonging to one class. The score function is the most unique part of the OCC algorithm. 

\subsubsection{OCC algorithms}
The existing score functions can be roughly categorized into boundary-based, distance-based, probability-based, fake-based, and subtask-based methods \citep{OCCreview}. 

The boundary-based method computes the boundary around a one-class. One-class Support Vector Machine \citep{[4]}, also known as Support Vector Data Description\citep{[5]}, considers the boundary around the one-class distribution. On the other hand, Isolation Forest \citep{[6]} considers the ensemble of random trees, which contain random boundaries.

The distance-based method computes distances or similarities from one class. The representation of one class is the average or the nearest training samples. The most straightforward solution is the distance from the average of the training samples. Moreover, Local Outlier Factor\citep{[7]} computes the similarity of local reachable density with the nearest samples. Average Localized Proximity considers the default hyperparameters \citep{[15]}.

A probability-based method computes the probability of belonging to one class. Gaussian Mixture Model \citep{[8]} computes the probability that the sample belongs to the Gaussian distribution created from one class.

The fake-based method \citep{[13]} creates pseudo samples belonging to other classes. Subsequently, the method trains a binary classification model between one class and fake classes.

The subtask-based method uses the error of subtasks, referring to other ML problems. The method trains arbitrary ML models to solve a subtask, using a training set containing only one class. Therefore, the model will produce smaller errors for samples in one class than those for other classes. The existing subtasks include reconstruction \citep{[9]}, transformation\citep{[10]}, knowledge distillation\citep{[11]} (also known as student-teacher\citep{[12]}), classification\citep{[22]}, and contrastive learning\citep{[14]}. 

Additionally, there is a method for combining multiple OCC algorithms/models. These studies are called ensemble learning or one-class ensemble \citep{[20],[21]}. Ensemble learning can improve classification performance compared to a single algorithm. 

Apart from these studies, this study aims to classify OCC models. This paper focuses solely on the models trained on feature vectors.

\subsection{Meta Learning and Explainable Artificial Intelligence}
\label{sec2.2}
This paper is related to meta-learning and explainable artificial intelligence. Meta learning \citep{metareview} is a technique for learning to learn. The goal is to improve the learning process with the experience of multiple learning episodes. The topic is related to hyperparameter tuning. 

Vanschoren \citep{metareview2} categorized meta-learning into three types. The first type is learning from model evaluation. The goal is to predict suitable hyperparameters from the evaluation metric and task. The second group is learning from the task property. The goal is to predict the evaluation from the hyperparameter and task. The third group is learning from prior models. The idea is to use prior models as the starting point of new models. The typical example is transfer learning \citep{transfer}, which fine-tunes the pre-trained models for new tasks. The classification of OCC models falls into the third category. Apart from meta-learning, the goal is to classify prior models. 

The classification of OCC models falls into the third category. Apart from meta-learning, the goal is to classify prior models. Yang et al. \citep{fed-poison} proposed a relevant method to detect poisoned models in federated learning. The idea was to detect abnormal gradient norms across models. Their methods analyzed the dynamic training process.  Moreover, the applicability was limited to the models having weights, such as neural network. On the other hand, this paper classifies pre-trained OCC models.

Explainable Artificial Intelligence (XAI)\citep{XAIreview} is a technique to explain the behavior of ML models as humans can understand. The XAI is categorized into three types \citep{XAIreview2}. The first type trains white box models, such as decision trees \citep{EDT}. The second type analyzes the black box models by perturbing inputs \citep{SHAP}. The third type is surrogating black box models by white box models \citep{LIME}. The classification of OCC models falls into the second type. The goal is to classify unlabeled OCC models by normality rankings. Classification of models can serve as a new type of explanation in XAI.

\subsection{Ranking correlation}
\label{sec2.3}
Ranking correlation is a similarity metric for rankings. This paper applies Spearman correlation \citep{[2]} and Kendall Tau correlation\citep{[3]} for normality rankings created by OCC models. Spearman correlation \citep{[2]} computes the distance between ranks of single items. 
\begin{equation}\label{eq5}
\tau = 1 - \frac{6\sum d_i^2}{n(n^2-1)}
\end{equation}
where $d_i$ is the distance of ranks for sample i. Moreover, n is the number of samples. 

On the other hand, Kendall Tau correlation \citep{[3]} considers the agreement regarding the orders of two items.
\begin{equation}\label{eq6}
\tau = \frac{P-Q}{\sqrt{(P+Q+T)(P+Q+U)}} 
\end{equation}
where P is the number of agreements, Q is the number of disagreements, and T and U are the variables to handle ties. n samples will create n(n-1) pairs. 

Spearman correlation is a faster variant, while the Kendall Tau correlation is a better alternative to handle ties.

\section{Classification algorithm of OCC models}
\label{sec3}
This section proposes classification algorithm for OCC models. The goal is to classify OCC models, as shown in equation (\ref{eq7}).

\begin{equation}\label{eq7}
Classification: model \rightarrow class
\end{equation}

The problem is similar to traditional supervised classification, consisting of training and testing stages. The training stage creates a classification model from the set of OCC models, while the testing stage applies the model to classify OCC models.

As described in subsection \ref{sec2.1}, the OCC model contains a classification function, a score function, and a threshold value. All of them can represent OCC models. This study analyzes score functions because it is the most unique part of OCC algorithms/models.

\begin{figure}[H]
\centering
\includegraphics[width=1\linewidth]{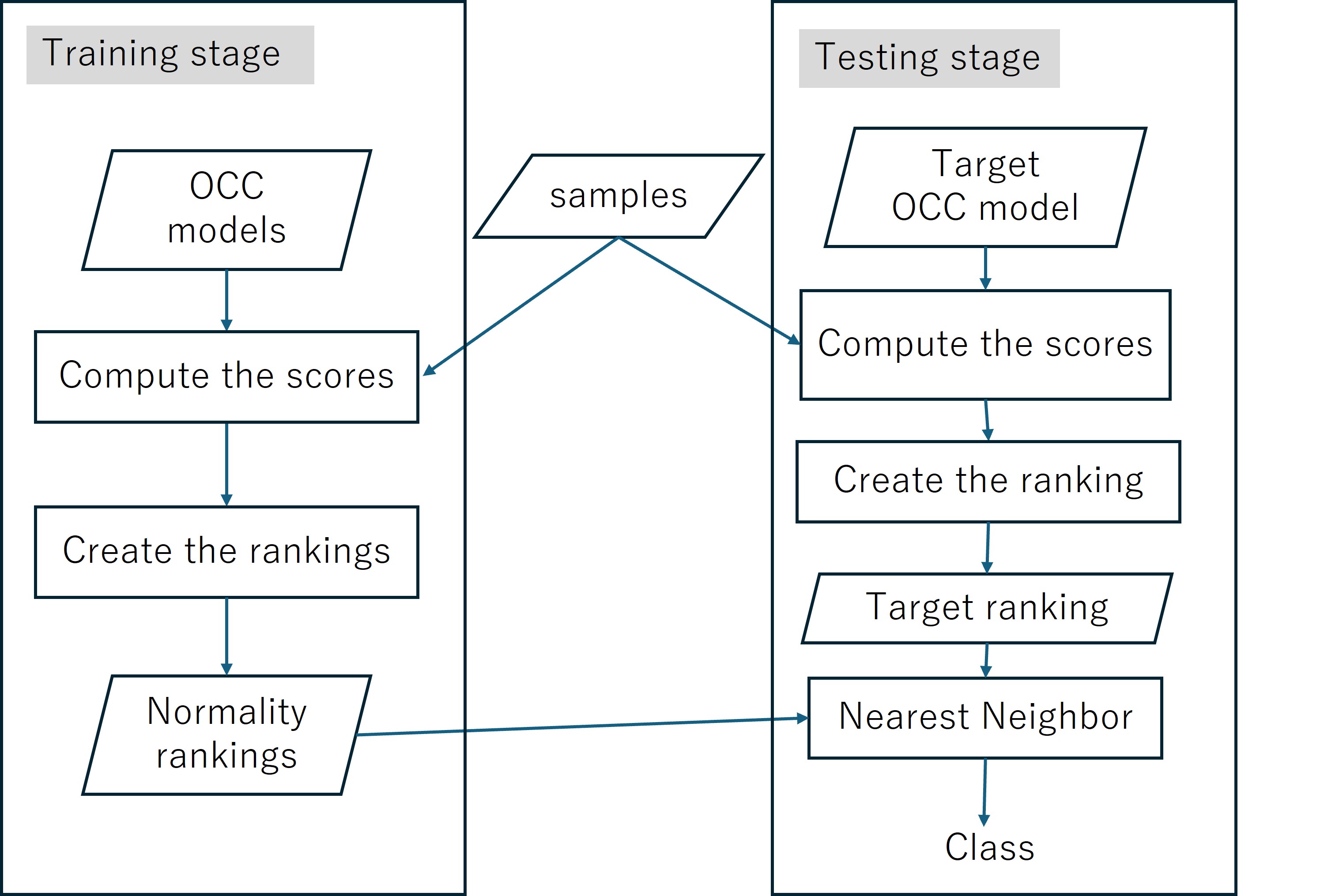}
\caption{Classification of OCC models}\label{fig1}
\end{figure}

Figure \ref{fig1} shows the proposed framework. The main idea is to create normality rankings from outputs of score functions. For this purpose, the algorithm prepares independent samples, called a ranking set. Subsequently, the ranking is classified by the nearest neighbor. 

\subsection{Represent OCC models as normality rankings}
\label{sec3.1}
This subsection represents the OCC algorithms as ranking as shown in Algorithm \ref{alg1}. The algorithm requires a ranking set, which contains samples to create the rankings. 

\begin{algorithm}[H]
\caption{Represent OCC models as rankings}\label{alg1}
\textbf{Input:} OCC models, ranking\_set

\textbf{Output:} Normality rankings

Scores=[]

\textbf{For} model in models:

\quad   score = model.score(ranking\_set) //Compute scores.
    
\quad  Scores.append(score)  

Scores=np.array(Scores)

Ranking=Scores.argsort(axis=1).argsort(axis=1) //Create normality ranking.

\end{algorithm}

\subsection{Supervised classification of ranking}
\label{sec3.2}
Subsequently, rankings are classified by the nearest neighbor. Ranking correlation metrics mentioned in section 2.2 are utilized to compute the similarity between rankings. The nearest neighbor is known as a lazy learner \citep{NN}, where the training stage does not exist. On the other hand, Algorithm \ref{alg3} shows the testing stage.

\begin{algorithm}[H]
\caption{Testing stage (Nearest Neighbor)}\label{alg3}
\textbf{Input:} rank\_test

\textbf{Given information:} rank\_train, y\_train (label)

\textbf{Output:} Classification result y\_pred.

Similarity\_OVO=[] //matrix (OVO: one vs one)

\textbf{For} rank1 in rank\_test:

\quad similarity=[]

\quad \textbf{For} rank2 in rank\_train:

\quad \quad similarity.append( rank\_correlation (rank1, rank2))

\quad Similarity\_OVO.append(similarity)

Similarity\_OVO = np.array(Similarity\_OVO) //

y\_pred = y\_train[Similarity\_OVO.argmax(axis=1)]

\end{algorithm}

\section{Experiment}
\label{sec4}
This section classifies the OCC models. Subsection \ref{sec4.1} describes the dataset and problem setting. Subsection \ref{sec4.2} shows the classification of the training dataset. Subsection \ref{sec4.3} summarizes the classification of OCC algorithms. Subsection \ref{sec4.4} reports the classification of hyperparameters.

\subsection{The dataset}\label{sec4.1}
This paper considers the classification of OCC models. Therefore, the dataset is the OCC models. The experiment applies four OCC algorithms, including one-class support vector machine (OCSVM)\citep{[4]}, Local Outlier Factor (LOF)\citep{[7]}, Isolation Forest (IF)\citep{[6]}, and Gaussian Mixture Model (GMM)\citep{[8]}. Moreover, its ensemble learning models are included in the classification target. The algorithms are implementation from the sklearn package \citep{sklearn} in Python as shown in Table \ref{table1}.

\begin{table}[H]
\caption{OCC algorithms from sklearn package\citep{sklearn}}\label{table1}
\centering
\scalebox{0.8}{
\begin{tabular}{l l l} \hline
Algorithms & Package & Hyperparameter \\ \hline
OCSVM \citep{[4]} & sklearn.svm.OneClassSVM & Default \\
LOF \citep{[7]} & sklearn.neighbors.LocalOutlierFactor & Novelty=True \\
IF \citep{[6]} & sklearn.ensemble.IsolationForest & Default \\
GMM \citep{[8]} & sklearn.mixture.GaussianMixture & Default \\
\hline
\end{tabular}
}
\end{table}

The experiment trains OCC models on the KDD cup dataset \citep{KDDCUP}. The dataset contains 97,278 samples and 396,747 abnormal samples. Note that the abnormal samples are separated into various sub-classes. However, the experiment gathers all abnormal samples into a single abnormal class for simplicity.

Before experimenting, it is worth noting that some OCC algorithms produce identical models when the training dataset and hyperparameters are identical. This aspect is useful for estimating whether the experiment settings are identical. However, the experiment will be boring if most OCC models are the same. 

\begin{figure}[H]
\centering
\includegraphics[width=1\linewidth]{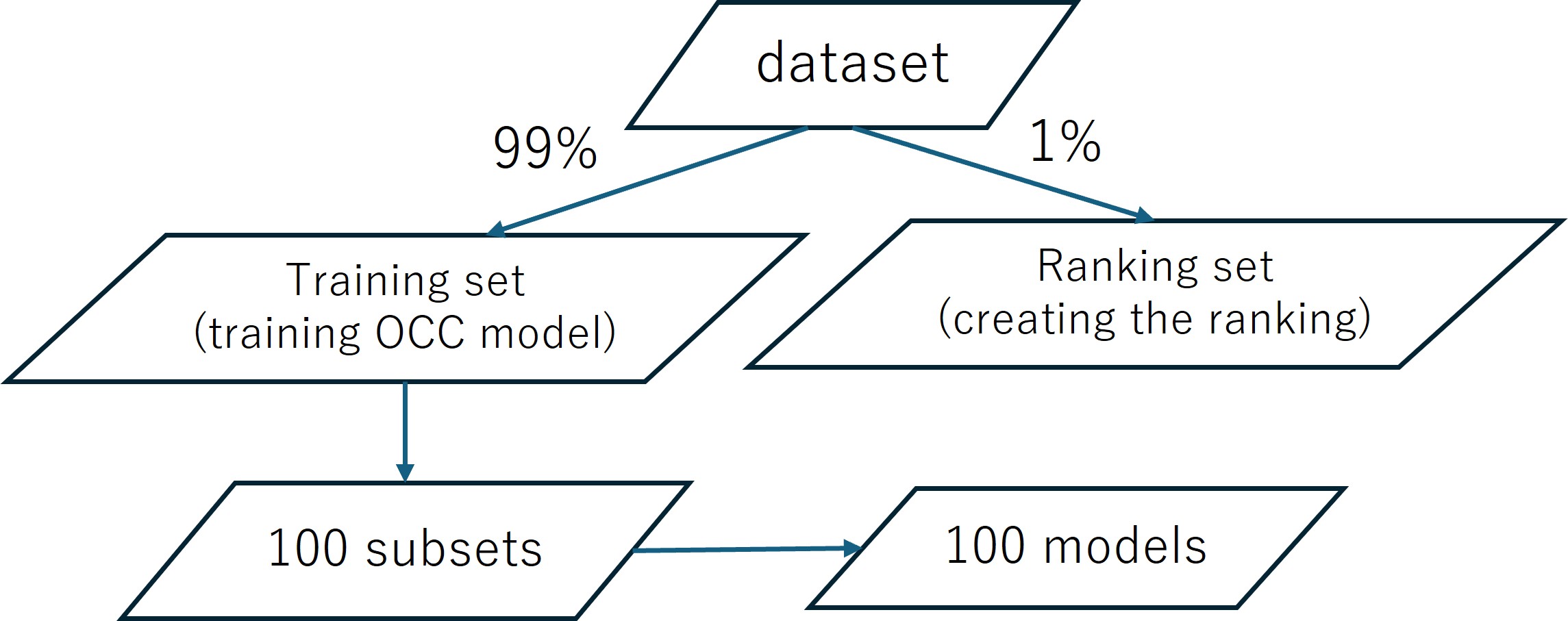}
\caption{The experiment procedure to train OCC models}\label{fig2}
\end{figure}

For this reason, the experiment separated the dataset into subsets as shown in Figure \ref{fig2}. First, the dataset is split into a training set and a ranking set with a 99:1 ratio. Subsequently, the training set is divided into 100 subsets to train 100 OCC models. This procedure is repeated for each class of OCC models.

The experiment considers the following classification problems.
\begin{itemize}
\item Subsection \ref{sec4.2}: Classification of training dataset (normal vs abnormal vs mix)
\item Subsection \ref{sec4.3}: Classification of OCC algorithms (OCSVM, LOF, IF, GMM, and ensemble) 
\item Subsection \ref{sec4.4}: Classification of hyperparameters. 
\end{itemize}

In addition, the experiment uses Spearman correlation for computing the similarity of rankings because it is a faster variant than Kendall Tau. Processing time is compared in subsection \ref{sec5.3}

\subsection{Classification of dataset}\label{sec4.2}
This subsection classifies OCC models trained on the different dataset characteristics. The first experiment is a binary classification between the models that learned the normal class and the models that learned the abnormal class. OCC models are trained from the KDD cup datasets \citep{KDDCUP}. For this purpose, 100 subsets are prepared as described in Figure \ref{fig2}. The experiment considered normal and abnormal classes separately. In particular, normal subsets contain 963 samples, while abnormal subsets contain 3927 samples. Moreover, the ranking set contains 4941 samples (973 normal and 3968 abnormal samples). The experiment ignored a few remaining samples.

Table \ref{table2} reports the accuracy of classifying the training dataset. The first experiment classifies the models that learned the normal class and those that learned the abnormal classes. The second experiment adds another class, namely mix, which contains both training and abnormal samples. The final task classifies the normal-abnormal ratios. Table \ref{table4} describes the details of 11 class labels.

\begin{table}[H]
\caption{Accuracy to classify training dataset (\%)}\label{table2}
\centering
{
\begin{tabular}{| c| c| r r|} \hline
 & &\multicolumn{2}{c|}{Cross-validation}  \\
Classification Task & |class| & leave-one-out & 2-fold \\ \hline
normal (963) vs abnormal (3927) & 2 & 100.0 & 100.0 \\
normal vs abnormal vs mix(963+3927) & 3 & 99.3 & 99.4 \\ 
Normal-abnormal ratio (see Table \ref{table4}) & 11 & 83.6 & 87.3 \\
\hline
\end{tabular}
}
\end{table}

Table \ref{table3} reports the confusion matrix for normal vs abnormal vs mix (leave-one-out cross-validation). Normal classes are classified correctly. However, six models from abnormal sets, are misclassified into mix, while three mix models are misclassified as abnormal.

\begin{table}[H]
\caption{Confusion matrix of dataset (\%)}\label{table3}
\centering
{
\begin{tabular}{|l|l|r r r|} \hline
 & & \multicolumn{3}{c|}{Actual} \\ \hline
 & & normal & abnormal & mix \\ \hline
& normal & 400 & 0 & 0 \\
Predicted & abnormal & 0 & 397 & 6 \\
& mix & 0 & 3 & 394 \\ \hline
\end{tabular}
}
\end{table}

Table \ref{table4} shows the list of classes for normal abnormal ratios, and class balances for normal and abnormal classes. The class balances are adjusted from the subsets created by Figure \ref{fig2}.
\begin{table}[H]
\caption{Class labels: normal-abnormal ratio}\label{table4}
\centering
{
\begin{tabular}{ l| l | r r}  \hline
Class & Ratio of normal/abnormal & normal & abnormal \\ \hline
0 & 100\% normal & 963 & 0 \\ 
1 & 90\% normal + 10\% abnormal & 963 & 107 \\
2 & 80\% normal + 20\% abnormal & 960 & 240 \\
3 & 70\% normal + 30\% abnormal & 959 & 411 \\
4 & 60\% normal + 40\% abnormal & 963 & 642 \\
5 & 50\% normal + 50\% abnormal & 963 & 963 \\
6 & 40\% normal + 60\% abnormal & 962 & 1443 \\
7 & 30\% normal + 70\% abnormal & 963 & 2247 \\
8 & 20\% normal + 80\% abnormal & 963 & 3852 \\
9 & 10\% normal + 90\% abnormal & 436 & 3924 \\
10& 100\% abnormal & 0 & 3927 \\ \hline
\end{tabular}
}
\end{table}
Table \ref{table5} shows the confusion matrix to classify normal abnormal ratio (leave-one-out). The model learned 100\% normal samples are classified with perfect accuracy, while classification error is observed when OCC models learn abnormal samples. This result suggests that learning abnormal samples change the OCC model significantly, while learning normal samples do not make the difference to the model.

\begin{table}[H]
\caption{Confusion matrix: classification of normal-abnormal ratio}\label{table5}
\centering
{
\begin{tabular}{ l| l | r r r r r r r r r r r}  \hline
 &  & \multicolumn{11}{c}{Actual}   \\ \hline
 &  & 0 & 1 & 2 & 3 & 4 & 5 & 6 & 7 & 8 & 9 & 10 \\ \hline
 & 0 & 400 & 0 & 0 & 0 & 0 & 0 & 0 & 0 & 0 & 0 & 0 \\
 & 1 & 0 & 396 & 4 & 0 & 0 & 0 & 0 & 0 & 0 & 0 & 0 \\
 & 2 & 0 & 4 & 385 & 8 & 0 & 0 & 0 & 0 & 0 & 0 & 0 \\
 & 3 & 0 & 0 & 11 & 339 & 42 & 0 & 0 & 0 & 0 & 0 & 0 \\
 & 4 & 0 & 0 & 0 & 52 & 290 & 55 & 0 & 0 & 0 & 0 & 0 \\
Predicted & 5 & 0 & 0 & 0 & 1 & 68 & 289 & 39 & 0 & 0 & 0 & 0 \\
 & 6 & 0 & 0 & 0 & 0 & 0 & 56 & 286 & 61 & 0 & 0 & 0 \\
 & 7 & 0 & 0 & 0 & 0 & 0 & 0 & 75 & 293 & 64 & 0 & 0 \\
 & 8 & 0 & 0 & 0 & 0 & 0 & 0 & 0 & 46 & 299 & 94 & 1 \\
 & 9 & 0 & 0 & 0 & 0 & 0 & 0 & 0 & 0 & 37 & 305 & 1 \\
 & 10 & 0 & 0 & 0 & 0 & 0 & 0 & 0 & 0 & 0 & 1 & 398 \\ \hline
\end{tabular}
}
\end{table}

Overall, the proposed method classified the characteristics of training datasets with high accuracy. Although classification of normal-abnormal ratio is 83.6\%, such a problem does not exist in the real-world. The proposed method has enough accuracy to classify the training dataset of OCC models.

\subsection{Classification of OCC algorithms}\label{sec4.3}
This section classifies the OCC models based on the algorithms. Moreover, the experiment included ensemble models as classification target. The ensemble process is voting, where all base learners are trained from the same subset. Ensemble score is the summation of normalized scores of base learners \citep{[30]}.
\begin{equation}\label{eq8}
score_{ens}(X) = \sum normalized(score_{base}(X))
\end{equation}
where the normalization is min-max, using the maximum and the minimum scores for the ranking set.

Table \ref{table6} shows the list of classes. There are 15 classes in total. 4 classes (0-3) are base learners while 6 classes (4-9) indicate the ensemble of two base learners. Moreover, 4 classes (10-13) combine three base learners. Finally, class 14 combines four algorithms.

\begin{table}[H]
\caption{Class labels of the models}\label{table6}
\centering
\begin{tabular}{l l} \hline
Class & OCC Algorithm \\ \hline
0 & OCSVM \\
1 & LOF \\
2 & IF \\
3 & GMM \\
4 & OCSVM+LOF \\
5 & OCSVM+IF \\
6 & OCSVM+GMM \\
7 & LOF+IF \\
8 & LOF+GMM \\
9 & IF+GMM \\
10 & OCSVM+LOF+IF \\
11 & OCSVM+LOF+GMM \\
12 & OCSVM+IF+GMM \\
13 & LOF+IF+GMM \\
14 & All \\ \hline
\end{tabular}

\end{table}

Table \ref{table7} reports classification accuracy for two tasks, 1) classification of base learners (4 classes) and 2) classification of ensemble models (15 classes). Overall, the proposed method classified base learners with high accuracy. However, the accuracy is degraded for classification of ensemble models. The result suggests that base learners have enough differences to be classified, while the ensemble models are relatively similar.

\begin{table}[H]
\caption{Classification Accuracy to distinguish OCC algorithms and one-class ensemble models}\label{table7}
\centering
{
\begin{tabular}{l|l| r r} \hline
 &  & \multicolumn{2}{c}{Cross-validation}  \\
Dataset & Classification target & leave-one-out & 2-fold \\ \hline
Normal & Base learners (4 classes) & 100.0 & 100.0 \\
KDD & Ensemble models(15 classes) & 85.5 & 84.3 \\ \hline
Abnormal & Base learners (4 classes) & 100.0 & 100.0 \\
KDD & Ensemble models(15 classes) & 23.7 & 34.1 \\ \hline
\end{tabular}
}
\end{table}

Table \ref{table8} shows the confusion matrix for the classification of ensemble models (normal class). Generally, the proposed methods classified base learners with high accuracy. However, the classification accuracy degrade after multiple models are combined.

\begin{table}[H]
\caption{Confusion Matrix: multi-class classification of one-class ensemble models (normal training samples, leave-one-class out cross-validation)}\label{table8}
\centering
\scalebox{0.8}{
\begin{tabular}{r | r | r r r r r r r r r r r r r r r r} \hline
 &  &  \multicolumn{15}{c}{Actual}  \\ \hline
 &  & 0 & 1 & 2 & 3 & 4 & 5 & 6 & 7 & 8 & 9 & 10 & 11 & 12 & 13 & 14 \\ \hline
 & 0 & 97 & 0 & 0 & 0 & 0 & 4 & 0 & 0 & 0 & 0 & 0 & 0 & 0 & 0 & 0 \\
 & 1 & 0 & 93 & 0 & 0 & 0 & 0 & 0 & 0 & 7 & 0 & 0 & 0 & 0 & 0 & 0 \\
 & 2 & 0 & 0 & 100 & 0 & 0 & 0 & 0 & 0 & 0 & 0 & 0 & 0 & 0 & 0 & 0 \\
 & 3 & 0 & 0 & 0 & 100 & 1 & 0 & 0 & 0 & 0 & 0 & 0 & 1 & 0 & 0 & 0 \\
 & 4 & 3 & 0 & 0 & 0 & 99 & 0 & 0 & 1 & 0 & 0 & 1 & 0 & 0 & 0 & 0 \\
 & 5 & 0 & 0 & 0 & 0 & 0 & 74 & 0 & 0 & 0 & 0 & 25 & 0 & 2 & 0 & 0 \\
 & 6 & 0 & 0 & 0 & 0 & 0 & 0 & 99 & 0 & 0 & 0 & 0 & 3 & 1 & 0 & 0 \\
Predicted & 7 & 0 & 0 & 0 & 0 & 0 & 0 & 0 & 65 & 0 & 0 & 0 & 0 & 0 & 20 & 0 \\
 & 8 & 0 & 7 & 0 & 0 & 0 & 0 & 0 & 0 & 92 & 0 & 0 & 0 & 0 & 0 & 0 \\
 & 9 & 0 & 0 & 0 & 0 & 0 & 0 & 0 & 0 & 0 & 87 & 0 & 0 & 1 & 5 & 0 \\
 & 10 & 0 & 0 & 0 & 0 & 0 & 21 & 0 & 15 & 0 & 0 & 74 & 0 & 1 & 0 & 2 \\
 & 11 & 0 & 0 & 0 & 0 & 0 & 0 & 1 & 0 & 0 & 0 & 0 & 96 & 0 & 0 & 3 \\
 & 12 & 0 & 0 & 0 & 0 & 0 & 1 & 0 & 0 & 0 & 3 & 0 & 0 & 71 & 0 & 27 \\
 & 13 & 0 & 0 & 0 & 0 & 0 & 0 & 0 & 19 & 1 & 8 & 0 & 0 & 0 & 68 & 0 \\
 & 14 & 0 & 0 & 0 & 0 & 0 & 0 & 0 & 0 & 0 & 2 & 0 & 0 & 24 & 7 & 68 \\ \hline
\end{tabular}
}

\end{table}

\subsection{Classification of hyperparameters}\label{sec4.4}
This subsection classifies OCC models based on its hyperparameters. Table \ref{table9} reports classification accuracy. The experiment considered five problem settings. The first problem is classification of kernels in OCSVM. The second problem is classification of number of neighbors for LOF. The third problem is a classification of max features, which is number of features contributing to create random trees in IF. The fourth and fifth problems are classification of hyperparameters in GMM. The fourth task classifies covariance type, while fifth task classifies number of components.

\begin{table}[H]
\caption{Classification accuracy to distinguish hyperparameters (\%)}\label{table9}
\centering
{
\begin{tabular}{l| l|l| r r} \hline
& &  & \multicolumn{2}{c}{Cross-validation}  \\
Dataset & Algorithm & Classification target & leave-one-out & 2-fold \\ \hline
Normal & OCSVM & Kernels (4 classes) & 93.8 & 94.0 \\ 
KDD & LOF & Number of neighbors (3-11) & 2.8 & 4.9 \\ 
& IF & Max features(1-10) & 27.9 & 26.1 \\ 
& GMM & Covariance type (4 classes) & 75.0 & 62.0 \\
& & Number of components (1-10) & 25.7 & 27.9 \\ \hline
Abnormal & OCSVM & Kernels (4 classes) & 100.0 & 100.0 \\ 
KDD & LOF & Number of neighbors (3-11) & 0.6 & 2.7 \\ 
& IF & Max features(1-10) & 25.0 & 23.5 \\ 
& GMM & Covariance type (4 classes) & 75.0 & 63.0 \\
& & Number of components (1-10) & 49.9 & 50.8 \\ \hline
\end{tabular}
}
\end{table}

The following tables visualize the confusion matrices of OCC models learned normal class. Table \ref{table10} shows the confusion matrix for kernels. Polynomial and RBF kernels were classified perfectly. However, there were classification errors between linear and sigmoid kernels.





\begin{table}[H]
\caption{Confusion matrix (Training data: Normal class of KDD)}\label{table10}
\centering
{
\begin{tabular}{l | l | r r r r|} \hline
 &  & \multicolumn{4}{c|}{Actual}  \\ \hline
 &  & Linear  & Poly & RBF & Sigmoid \\ \hline
Predicted & Linear & 89 & 0 & 0 & 14 \\
 & Poly & 0 & 100 & 0 & 0 \\
 & RBF & 0 & 0 & 100 & 0 \\
 & Sigmoid & 11 & 0 & 0 & 86 \\ \hline
\end{tabular}
}
\end{table}

In addition, Table \ref{table11} shows the confusion matrix for number of neighbors of LOF. Although classification accuracy is 2.8\%, the prediction is similar to actual labels. This result is fair as a regression problem. 

\begin{table}[H]
\caption{Confusion matrix of number of neighbors for LOF}\label{table11}
\centering
{
\begin{tabular}{l | l | r r r r r r r r r|} \hline
 &  & \multicolumn{9}{c|}{Actual} \\ \hline
 &  & 3 & 4 & 5 & 6 & 7 & 8 & 9 & 10 & 11 \\ \hline
 & 3 & 5 & 34 & 0 & 0 & 0 & 0 & 0 & 0 & 0 \\
 & 4 & 70 & 1 & 22 & 0 & 0 & 0 & 0 & 0 & 0 \\
 & 5 & 14 & 56 & 2 & 33 & 1 & 0 & 1 & 0 & 0 \\
 & 6 & 7 & 3 & 63 & 3 & 32 & 2 & 1 & 1 & 1 \\
Predicted & 7 & 3 & 5 & 6 & 59 & 2 & 38 & 2 & 2 & 2 \\
 & 8 & 1 & 0 & 2 & 4 & 57 & 1 & 50 & 6 & 1 \\
 & 9 & 0 & 1 & 4 & 0 & 2 & 47 & 1 & 39 & 11 \\
 & 10 & 0 & 0 & 1 & 1 & 4 & 11 & 43 & 2 & 77 \\
 & 11 & 0 & 0 & 0 & 0 & 2 & 1 & 2 & 50 & 8 \\ \hline
\end{tabular}
}
\end{table}

Table \ref{table12} shows confusion matrix of max feature value for IF. Although the accuracy is 27.9\%, the classification result is fair as a regression.

\begin{table}[H]
\caption{Confusion matrix of max\_feature in IF}\label{table12}
\centering
{
\begin{tabular}{l | l | r r r r r r r r r r|} \hline
 &  & \multicolumn{10}{c|}{Actual} \\ \hline
 &  & 1 & 2 & 3 & 4 & 5 & 6 & 7 & 8 & 9 & 10 \\ \hline
 & 1 & 39 & 11 & 1 & 0 & 0 & 0 & 0 & 0 & 0 & 0 \\
 & 2 & 29 & 18 & 15 & 4 & 2 & 2 & 0 & 0 & 0 & 0 \\
 & 3 & 19 & 33 & 37 & 13 & 8 & 4 & 0 & 0 & 0 & 0 \\
 & 4 & 9 & 21 & 21 & 20 & 14 & 7 & 6 & 2 & 0 & 1 \\
Predicted & 5 & 2 & 9 & 16 & 19 & 24 & 18 & 11 & 8 & 8 & 3 \\
 & 6 & 0 & 4 & 6 & 19 & 18 & 24 & 9 & 6 & 11 & 4 \\
 & 7 & 2 & 2 & 2 & 19 & 14 & 25 & 27 & 21 & 9 & 12 \\
 & 8 & 0 & 2 & 1 & 2 & 11 & 9 & 26 & 23 & 17 & 20 \\
 & 9 & 0 & 0 & 1 & 2 & 7 & 5 & 8 & 23 & 23 & 16 \\
 & 10 & 0 & 0 & 0 & 2 & 2 & 6 & 13 & 17 & 32 & 44 \\ \hline
\end{tabular}
}
\end{table}

Table \ref{table13} shows confusion matrix of covariance type of GMM. There are four alternatives \citep{sklearn}.
\begin{itemize}
\item Full: each component has its own general covariance matrix.
\item Tied: all components share the same general covariance matrix.
\item Diag: each component has its own diagonal covariance matrix.
\item Spherical: each component has its own single variance.
\end{itemize}
Indeed, the experiment set the number of component as 1, where full and tied are identical, resulting the mis-classification between them.

\begin{table}[H]
\caption{Classification of covariance types for GMM (number of components = 1).}\label{table13}
\centering
{
\begin{tabular}{l | l | r r r r|} \hline
 &  & \multicolumn{4}{c|}{Actual}  \\ \hline
 &  & Full & Tied & Diag & Spherical \\ \hline
Predicted & Full & 46 & 46 & 0 & 0 \\
 & Tied & 54 & 54 & 0 & 0 \\
 & Diag & 0 & 0 & 100 & 0 \\
 & Spherical & 0 & 0 & 0 & 100 \\ \hline
\end{tabular}
}
\end{table}

Table \ref{table14} shows the confusion matrix of GMM regarding number of Gaussian Mixtures. Classification is successful, when the number of component is 1, while the result degrades by increasing Gaussian Mixtures. Perhaps, increasing Gaussian Mixture does not make the difference. 

\begin{table}[H]
\caption{Confusion matrix for number of components in GMM (covariance = full)}\label{table14}
\centering
{
\begin{tabular}{l | l | r r r r r r r r r r|} \hline
 &  & \multicolumn{10}{c|}{Actual} \\ \hline
 &  & 1 & 2 & 3 & 4 & 5 & 6 & 7 & 8 & 9 & 10 \\ \hline
 & 1 & 93 & 18 & 5 & 1 & 0 & 0 & 1 & 0 & 0 & 0 \\
 & 2 & 4 & 49 & 15 & 10 & 2 & 0 & 1 & 0 & 0 & 0 \\
 & 3 & 2 & 17 & 39 & 23 & 7 & 7 & 1 & 1 & 1 & 1 \\
 & 4 & 1 & 7 & 20 & 20 & 18 & 5 & 5 & 1 & 4 & 3 \\
Predicted & 5 & 0 & 5 & 10 & 18 & 18 & 26 & 8 & 2 & 3 & 6 \\
 & 6 & 0 & 1 & 2 & 9 & 31 & 13 & 27 & 13 & 6 & 12 \\
 & 7 & 0 & 1 & 4 & 3 & 9 & 23 & 6 & 24 & 12 & 7 \\
 & 8 & 0 & 1 & 1 & 5 & 9 & 14 & 27 & 5 & 40 & 30 \\
 & 9 & 0 & 0 & 1 & 3 & 2 & 4 & 17 & 35 & 7 & 34 \\
 & 10 & 0 & 1 & 3 & 8 & 4 & 8 & 7 & 19 & 27 & 7 \\ \hline
\end{tabular}
}
\end{table}

The hyperparameters can be roughly grouped into categorical (kernel and covariance type) and numerical (others). Classification accuracy is high for categorical hyperparameter. On the other hand, numerical hyperparameter is not classified with high accuracy. This result suggests that minor change of numerical hyperparameters do not create significant difference to the models. The classification result is fair as a regression problem. 

\section{Discussion}\label{sec5}
This section discusses several aspects of the proposed method. Subsection\ref{sec5.1} states the advantages and limitations. Subsection \ref{sec5.2} applies the proposed method to analyze sliding windows of biometric signal. Subsection \ref{sec5.3} demonstrates classification of datasets using sleep records. Subsection \ref{sec5.4} is a time complexity analysis, while subsection \ref{sec5.5} conducts an ablation study for the ranking set. Finally, subsection \ref{sec5.6} discusses the classification of other ML models.

\subsection{Advantage and limitation}\label{sec5.1}
This paper presents new findings: classification of OCC models is possible with high accuracy. Roughly speaking, the OCC model is a representation of the dataset. Therefore, the classification of OCC models is essentially the classification of datasets. Perhaps classification of the dataset is easier than traditional classification to samples because datasets have more differences than single samples.

Classification of OCC models could tackle several problems. The first application is a problem of combining multiple ML models, such as ensemble learning and federated learning \citep{federatedlearning}. In particular, federated learning is a promising application because it aims to keep training samples private. The results in this paper can detect situations where a private training set differs from those of other machines.

Another application is the classification of datasets. The traditional classification processed samples one by one. On the other hand, classification of datasets can classify multiple samples together, potentially leading to new research directions. Similarly, the part of this paper is applicable to classify the rankings. 

The proposed method provides a unified solution for classifying the dataset, the OCC model, and ranking. Moreover, one can represent single samples as datasets, which can be represented as OCC models and rankings. For example, sliding windows can create a dataset from a sample.

On the other hand, the paper has several limitations. The first limitation is processing speed to compute the nearest neighbors, which has scalability issue for large scale rankings. The second issue is a gap with real world. The experiment in this paper had access to score functions. However, the actual environment might provide only classification function. Moreover, creating ranking set will be challenge for analyzing black-box models, where input shape is unknown.

\subsection{The potential extension in healthcare}\label{sec5.2}
This section demonstrates the proposal in the healthcare domain. For this purpose, we use a breathing disorder dataset \citep{[33]} created by the University of Hradec Králové. The dataset contains Ballistocardiogram (BCG), a signal caused by the heart vibration.
The signal is collected from participants who follow the scheduled breathing behaviors(see detail in \citep{[34]}), which include breath holding and body position changes. The experiment uses the signal for participant ID 1.

\begin{figure}[H]
\centering
\includegraphics[width=1\linewidth]{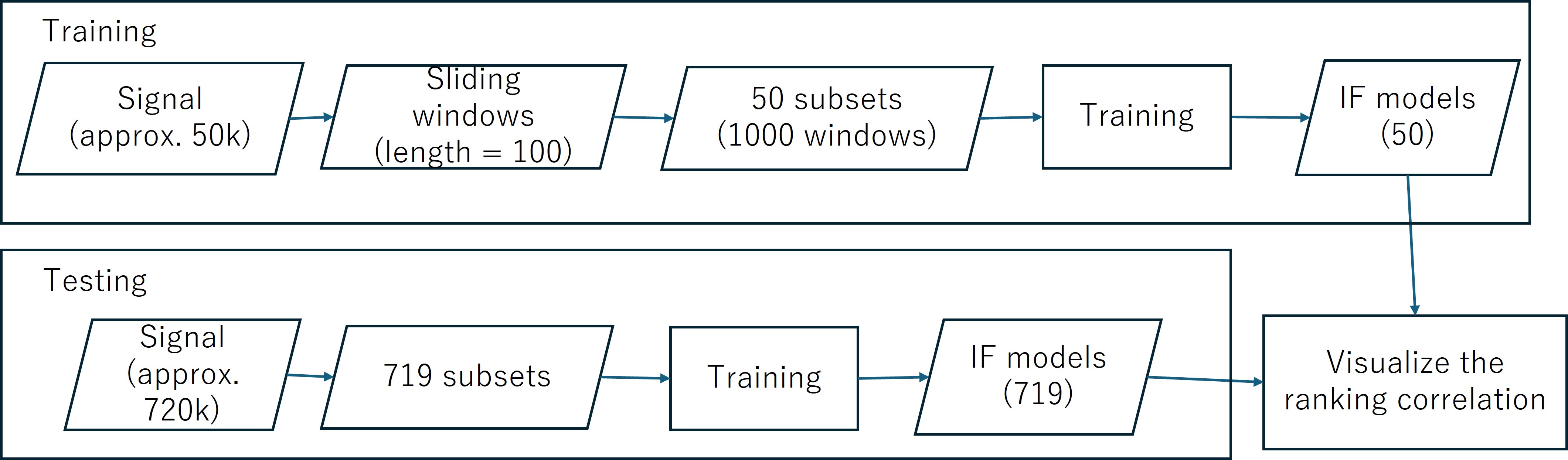}
\caption{Experiment procedure of breathing disorder dataset}\label{fig3}
\end{figure}

Figure \ref{fig3} shows the preprocessing image. The signal contains approximately 720k indices that are preprocessed into 100-length sliding windows. The training part is approximately 50k index, separated into 50 subsets, where each subset contains 1000 windows. The testing part consists of the entire index, divided into 719 subsets containing 1000 windows. Note that the first 50 subsets are identical to the training set. The experiment selected the IF model because it is the fastest OCC algorithm. Moreover, sliding windows in training parts (50k windows) are utilized as a ranking set.

\begin{figure}[H]
\centering
\includegraphics[width=1\linewidth]{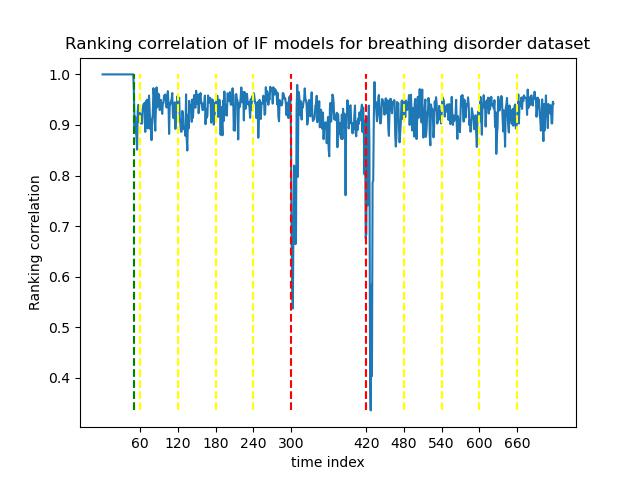}
\caption{Analyzing breathing disorder using the classification of OCC models}\label{fig4}
\end{figure}

Figure \ref{fig4} shows the Spearman correlation to the most similar ranking. The training set is before the green line. The participants hold their breath at the yellow lines, while they change their body position at the red line. Ranking correlation decreased around the red line.

Accordingly, the classification of OCC models could contribute to biometric signal analysis. The core idea is to train sliding windows by OCC models and classify these models. The details should be discussed in future work.

\subsection{Classification of dataset}\label{sec5.3}
Datasets and OCC models have one-to-one correspondence because OCC models are representations of datasets. Apart from the traditional single-sample classification, the classification of datasets distinguishes multiple samples. This subsection demonstrates the function using sleep records.

Biometric signals were measured and annotated at project Healthy Aging in Industrial Environment CZ.02.1.01/0.0/0.0/16\_019/0000798 \citep{HAIE}
 using HealthReact system \citep{Healthreact}, which is a cloud-based solution for data gathering, evaluation and reacting based on given rules. Subsequently, the activities are summarized at minute, hour, and daily-levels. Dataset contains large amount of data. This study use minute-level summaries that include user ID, sleep (true or false), active level (degree of movement), number of steps, and heart rate. Subsequently, this study preprocessed the minute-level summary into sleep records that include the start and end times of sleep and the duration. Continuous minutes with sleep = true were treated as a single sleep record.

This subsection demonstrates dataset classification through a case study of sleep records collected from 1308 participants. Each participant has a different number of sleep records, ranging from 1 to 570 (some participants stopped measuring after a few sleep records, while others continued for a year). The data analysis is performed on 100 people with the largest sleep records ($\geq$ 428). The experiment treats a single person as a single dataset. However, there is no annotation for datasets. 

Accordingly, the experiment applied unsupervised outlier detection to determine abnormal(minority) sleep behavior. Unsupervised outlier detection is the problem to detect outliers without definitions of normal and abnormal. Generally, normal samples refer to the majority, while abnormal samples are a minority. 

\begin{figure}[H]
\centering
\includegraphics[width=1\linewidth]{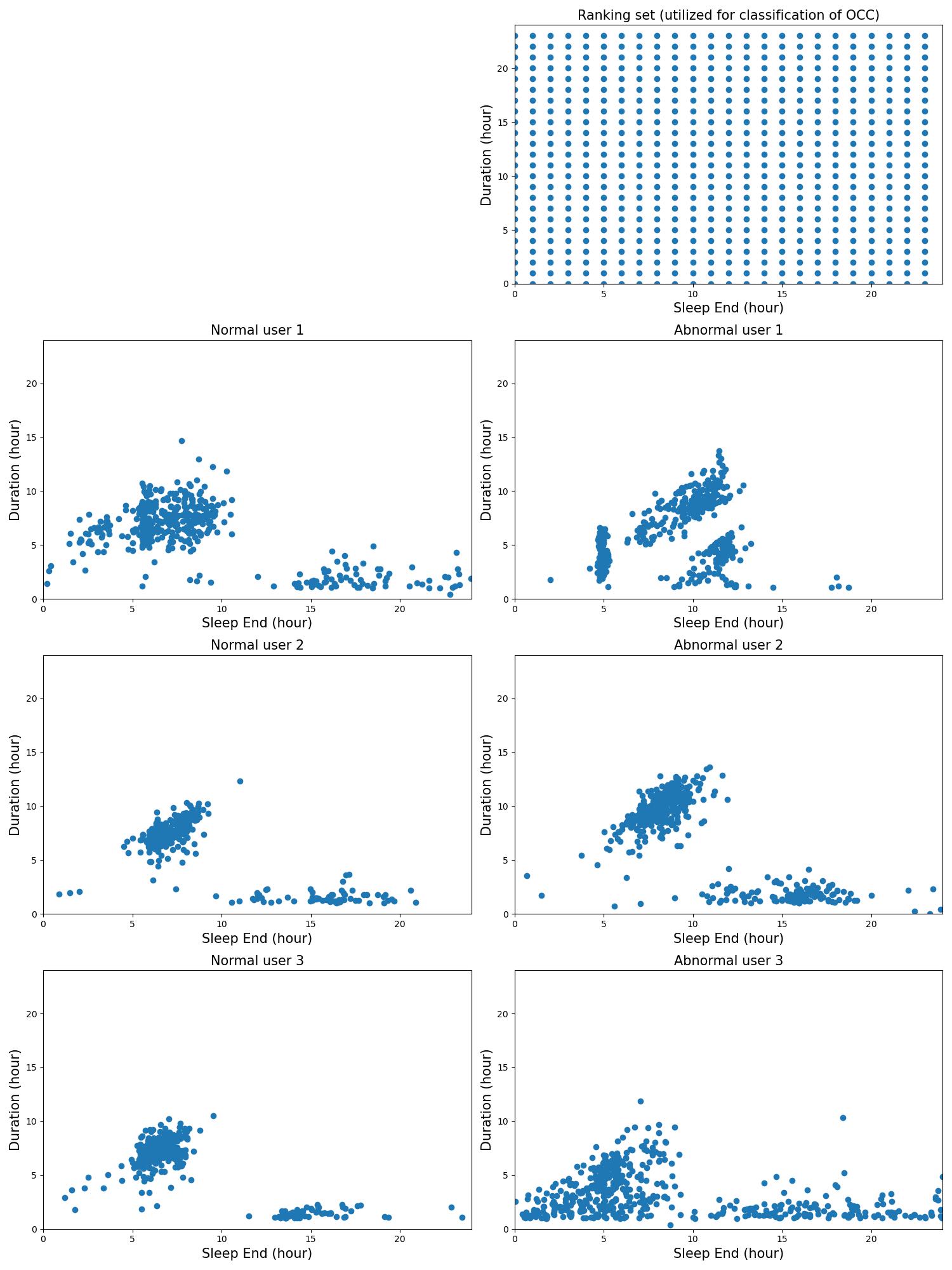}
\caption{Unsupervised outlier detection result using the classification of OCSVM models}\label{fig5}
\end{figure}

We trained OCSVM models on 100 users and computed normality rankings using a ranking set of 576 samples covering 24 $\times$ 24 integer hours. Figure \ref{fig5} visualizes the most normal and abnormal users (3 users for each). The normal score is the average ranking correlation to other users. The result demonstrates that the proposal has the potential to classify datasets.

Note that the result differs due to the choice of OCC algorithms. The hypotheses differ across OCC algorithms and they will create different normality rankings. Future work should discuss the choice of OCC algorithms and hyperparameters for classification.

\subsection{Time complexity analysis}\label{sec5.4}
The proposed approach shows high accuracy. However, it has scalability issues with large datasets. The main bottleneck is in computing the ranking correlations.

Spearman's Correlation is a faster alternative because it considers only the ranks of a single item. The computation cost is O($n$), where $n$ is the number of items. Kendall Tau correlation is time-consuming for considering the order of item pairs; the number of item pairs is approximately ($n^{2}$).

This time complexity must be applied to $N \times N$ ranking pairs. This paper compares the computation time required to rank correlations for 4400 models. The processing time includes applying OCC models to the ranking set and computing the ranking correlations. 

We applied the Scipy package \citep{[35]} to compute the Kendall Tau correlation, which has a reasonable processing speed. However, the Spearman correlation in the SciPy implementation \citep{[35]} is slow. The potential reason is that the package also computes the p-values. Accordingly, we created a faster implementation by computing only equation (\ref{eq5}) in a single line ($1-(((abs(rank1-rank2)**2).sum()*6)/(n**3-n))$), where rank1 and rank2 are numpy arrays.

\begin{table}[H]
\caption{Processing time to compute OVO correlations for 4400 rankings (4941 items)}\label{table15}
\centering
{
\begin{tabular}{l l l} \hline
Rank Correlation & Time &  \\
Kendall Tau \citep{[3]}(Scipy \citep{[35]}) & 3 h 38 min \\
Spearman \citep{[2]}(Scipy \citep{[35]}) * & 3 h 39 min  \\
Spearman \citep{[2]} & 3 min 46 s  \\ \hline
\end{tabular}
}
\end{table}

Currently, Spearman correlation is the only choice for large-scale models. Computing OVO correlations is time-consuming. Possible solutions include reducing the models or computing the ensemble models as a representation of classes.  

\subsection{Ranking set and ablation study}\label{sec5.5}
Theoretically, the proposed method can classify OCC models with high accuracy. However, the proposal must represent OCC models as normality rankings. The experiment separated the public dataset into a training set and a ranking set. However, the actual problems might classify unknown models. For this purpose, the first step is to identify the input shape of OCC models. Subsequently, the ranking set can be pseudo-samples with such a shape.

To confirm the above hypothesis, this subsection reports the ablation study of the ranking set. Table \ref{table16} compares four alternatives.
\begin{itemize}
\item Both: Ranking set contains both normal and abnormal classes (The results reported in the previous section)
\item Normal: Ranking set contains solely the normal class.
\item Abnormal: Ranking set contains solely the abnormal class.
\item Generated: Ranking set contains randomly generated samples. The size is the same as "Both". Each vector has random values in the range of 0 - 1. 
\end{itemize}

\begin{table}[H]
\caption{Ablation study of ranking set(leave-one-out cross-validation)}\label{table16}
\centering
{
\begin{tabular}{l|l| r r r r} \hline
 &  & \multicolumn{4}{c}{Ranking set}  \\
Dataset & Classification target & Both & Normal & Abnormal & Generated \\ \hline
Normal & Base learners (4 classes) & 100.0 & 100.0 & 100.0 & 100.0 \\
KDD & Ensemble models(15 classes) & 85.5 & 15.1 & 88.6 & 45.9 \\ \hline
Abnormal & Base learners (4 classes) & 100.0 & 100.0 & 100.0 & 100.0  \\
KDD & Ensemble models(15 classes) & 23.7 & 43.0 & 22.2 & 38.2 \\ \hline
Both & Normal vs Abnormal (2 classes) & 100.0 & 100.0 & 100.0 & 100.0 \\
 & Normal vs Abnormal vs Mix (3 classes) & 99.3 & 83.3 & 83.3 & 95.5 \\
 & Normal-abnormal ratio (11 classes) & 83.6 & 45.9 & 80.2 & 48.7 \\ \hline
\end{tabular}
}
\end{table}

Overall, the accuracy did not degrade for the simple problems (classification of base learners and normal vs abnormal), which is a promising result for future applications.

However, the classification accuracy degraded for the ensemble models and the normal-abnormal ratio. One important observation is that the classification of normal models degrades accuracy when the ranking set consists solely of the normal class. The same statement applies to the classification of abnormal models. Perhaps the ranking set should contain the samples that are not included in the training set of OCC models.

\subsection{Classification of other ML models}\label{sec5.6}
This paper leaves the classification of binary and multi-class models for future work, as these problems are complex. The solution must consider the pairs/groups of classes in the training dataset. In particular, creating rankings from multi-class classification models is challenging.

The potential solution is to approximate binary or multi-class classification models as a set of OCC models, where the class probabilities serve as the score functions. The idea is related to multi-view learning \citep{multi-view}, where each view approximates an OCC model. In addition, future work should classify further ML tasks, such as deep learning, and the models presented in this paper.

\subsubsection{Is it possible to classify the classification models for OCC models?}\label{sec5.6.1}
One possible extension is to classify the classification models for OCC models. To clarify, this subsection discusses models as layers. 

\begin{itemize}
\item The first layer is a traditional OCC model to classify samples.
\item The second-layer model classifies OCC models (discussed in this paper).
\item The third-layer model classifies the second-layer models.
\item There might be more layers to classify the previous layers.
\end{itemize}

Perhaps, developing the third-layer models is possible. However, preparing a ranking set that includes second-layer (or higher) models is challenging. Moreover, a key question is whether third-layer (or other) models can classify themselves, or endless layers are required to classify previous layers. Future work should address this question.

\section{Conclusion}
This paper proposed a classification algorithm for OCC models. The proposed method represented OCC models as normality rankings and classified the rankings by the nearest neighbor and ranking correlations. The proposed method classified OCC models from different training datasets with high accuracy. Moreover, the method can classify OCC algorithms when the training datasets contain the same class.

The proposed method is a unified solution to classify datasets, OCC models, and rankings. Moreover, the classification of OCC models could play an important role in classification of other ML models.

\section*{CRediT authorship contribution statement}
\textbf{Toshitaka Hayashi}: Conceptualization, Methodology, Writing – original draft, Investigation, Software, Visualization. \textbf{Hamido Fujita}: Writing – review and editing, Supervision.  \textbf{Dalibor Cimr}: Writing – original draft. \textbf{Richard Cimler}: Project administration, Funding acquisition. \textbf{Jitka Kuhnová}: Data curation.

\section*{Declaration of Competing Interest}
\label{decCI}
The authors declare that they have no known competing financial interests or personal relationships that could have appeared to influence the work reported in this paper. 

\section*{Acknowledgement}
\label{ack}
This study is supported by the research project “2200/04/2024-2026” as part of the "Competition for 2024-2026 Postdoctoral Job Positions at the University of Hradec Králové", at the Faculty of Science, University of Hradec Králové. This study is financially supported from the project "Research of Excellence on Digital Technologies and Wellbeing CZ.02.01.01/00/22\_008/0004583“, which is co-financed by the European Union. 

The data described in the study is from the project "Research of Excellence on Digital Technologies and Wellbeing“ (CZ.02.01.01/00/22\_008/0004583) which is co-financed by the European Union.

\bibliography{reference.bib}
\biboptions{authoryear}
\bibliographystyle{elsarticle-harv} 

\end{document}